%% file: main.tex
\documentclass[10pt,twocolumn,letterpaper]{article}

\usepackage[pagenumbers]{iccv} 
\usepackage{multirow}

\definecolor{iccvblue}{rgb}{0.21,0.49,0.74}
\usepackage[pagebackref,breaklinks,colorlinks,allcolors=iccvblue]{hyperref}

\def\paperID{5498} 
\def\confName{ICCV}
\def\confYear{2025}

\title{GenWorld: Towards Detecting AI-generated Real-world Simulation Videos}
\author{Weiliang Chen$^{2,}$ \quad Wenzhao Zheng$^{1,}$\footnotemark[1]\quad  Yu Zheng$^{1}$ \quad Lei Chen$^{1}$\quad Jie Zhou$^{1}$\quad Jiwen Lu$^{1}$\quad Yueqi Duan$^{2}$\footnotemark[2]\\
Department of Automation, Tsinghua University, China \\
Department of Electronic Engineering, Tsinghua University, China \\
\texttt{\small cwl24@mails.tsinghua.edu.cn; wenzhao.zheng@outlook.com;} \\
\texttt{\small \{yu-zheng, leichenthu, jzhou, lujiwen,duanyueqi\}@tsinghua.edu.cn}
}

\begin{document}
\twocolumn[{%
\renewcommand\twocolumn[1][]{#1}%
\maketitle
\vspace{-8mm}
\begin{center}
    \centering
    \includegraphics[width=\linewidth]{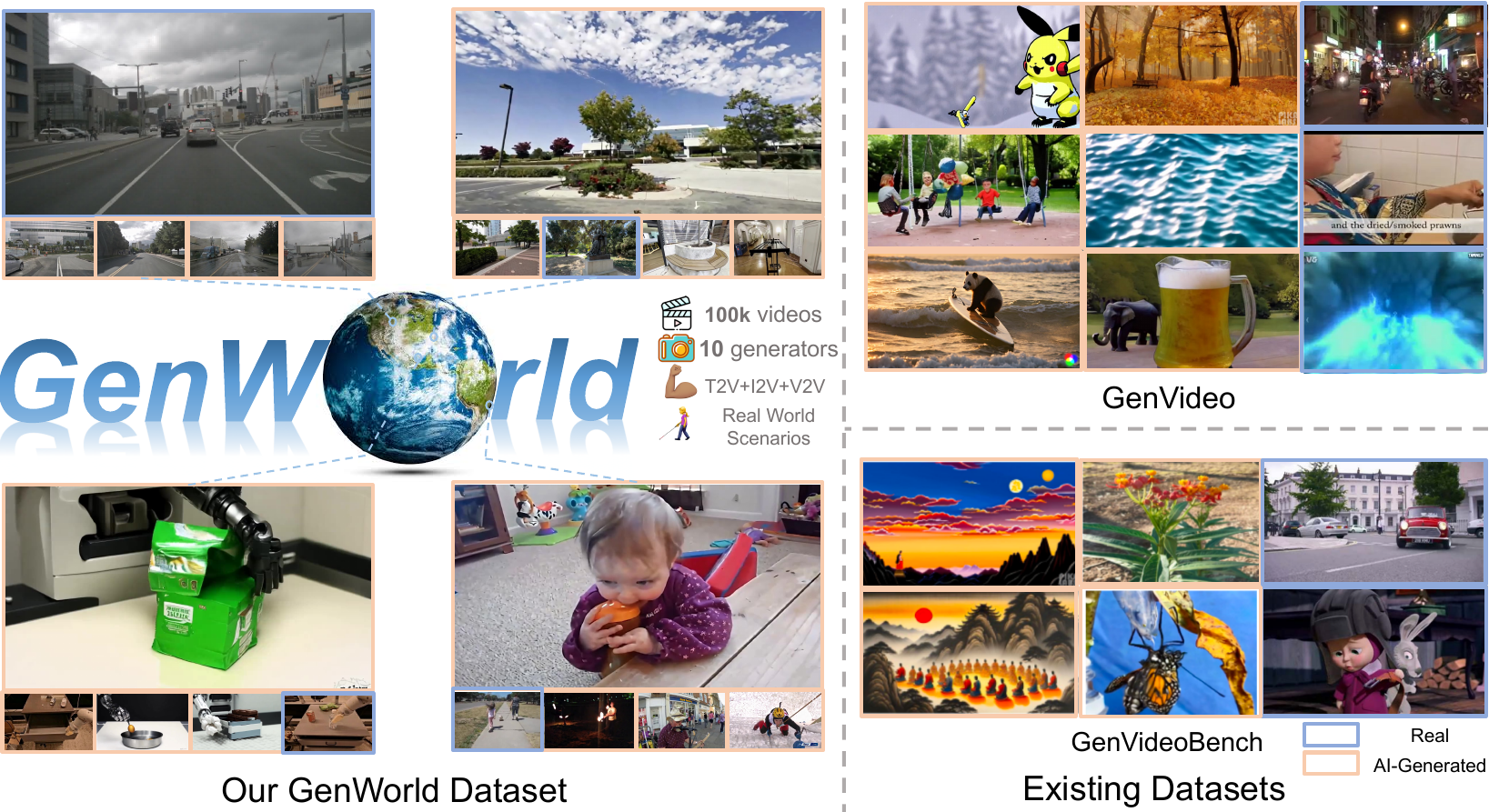}
    \vspace{-7mm}
    \captionof{figure}{
    Most existing AI-generated video datasets consist of cartoon videos even as ``real'' data, lacking a clear definition of authenticity.
    This paper proposes a high-quality dataset including only real and generated videos from real-world scenarios (e.g., driving, navigation, manipulation).
    \textbf{GenWorld} features three key characteristics: 1) \textbf{Real-world Simulation}, 2) \textbf{High Quality}, and 3) \textbf{Cross-prompt Diversity}, which can serve as a foundation for AI-generated video detection research with practical significance.
    }
\label{fig:teaser}
\vspace{1.2mm}
\end{center}
}]
\footnotetext[1]{Project Leader}
\footnotetext[2]{Corresponding Author}
\maketitle
\input{sec/0_abstract}    
\input{sec/1_intro}
\input{sec/2_related_work}
\input{sec/3_GenWorld}

\input{sec/4_SpannDetector}
\input{sec/5_experiments}
\input{sec/6_conclusion}

{
    \small
    \bibliographystyle{ieeenat_fullname}
    \bibliography{main}
}

\end{document}

%% file: sec/0_abstract.tex
\begin{abstract}
The flourishing of video generation technologies has endangered the credibility of real-world information and intensified the demand for AI-generated video detectors.
Despite some progress, the lack of high-quality real-world datasets hinders the development of trustworthy detectors. 
In this paper, we propose \textbf{GenWorld}, a large-scale, high-quality, and real-world simulation dataset for AI-generated video detection. 
GenWorld features the following characteristics: (1) \textbf{Real-world Simulation}: GenWorld focuses on videos that replicate real-world scenarios, which have a significant impact due to their realism and potential influence; (2) \textbf{High Quality}: GenWorld employs multiple state-of-the-art video generation models to provide realistic and high-quality forged videos; (3) \textbf{Cross-prompt Diversity}: GenWorld includes videos generated from diverse generators and various prompt modalities (e.g., text, image, video), offering the potential to learn more generalizable forensic features. 
We analyze existing methods and find they fail to detect high-quality videos generated by world models (i.e., Cosmos~\cite{agarwal2025cosmos}), revealing potential drawbacks of ignoring real-world clues.
To address this, we propose a simple yet effective model, SpannDetector, to leverage multi-view consistency as a strong criterion for real-world AI-generated video detection. 
Experiments show that our method achieves superior results, highlighting a promising direction for explainable AI-generated video detection based on physical plausibility. 
We believe that GenWorld will advance the field of AI-generated video detection.
Project Page: https://chen-wl20.github.io/GenWorld
\end{abstract}
\vspace{-5mm}

%% file: sec/1_intro.tex
\section{Introduction}
Recently, generative models~\cite{blattmann2023stablevideodiffusion, rombach2022highlatentdiffusion, zheng2024opensora} have been revolutionizing the world with their powerful capability to generate multimodal data, enabling a wide range of downstream applications~\cite{bar2024navigationworldmodel, agarwal2025cosmos, zheng2024occworld}. Among them, video generation models~\cite{zheng2024opensora, ma2024latte, chen2023seine, agarwal2025cosmos, zeroscopexl} have garnered the most attention, as video is the most expressive and information-rich modality for capturing the real world with a potential of serving as a foundation for world models~\cite{xiang2024pandora, huang2024owl1, videoworldsimulators2024sora}. 
Despite offering significant convenience, they have raised concerns about the authenticity of real-world information, highlighting the urgent need for effective AI-generated video detectors~\cite{barrett2023identifyingriskofgenerative, neupane2023impactsriskgenerative}. 

Despite some progress, the lack of high-quality real-world AI-generated video detection datasets has significantly limited the development of trustworthy detectors for real scenarios.
Early works~\cite{gu2021spatiotemporaldeepfake1method, xu2023talldeepfake2method, HCILdeepfake3method} have focused on curating deepfake detection datasets for face forgery detection, yet research beyond human faces is still limited. 
Pioneered by GVF~\cite{ma2024decofGVF}, several efforts~\cite{bai2024aiGVD, chen2024demamba, ni2025genvidbench} have started focusing on building general AI-generated video detection datasets. 
However, they prioritize large-scale data collection and ignore the quality and genres of generated videos, which are often disorganized and of inconsistent quality.
As shown on the right side of Figure~\ref{fig:teaser}, most existing datasets consist of cartoon videos even for ``real'' data.
This makes detectors trained on them to focus on generation flaws instead of physical plausibility, which is more important with rapid improvements in video generation quality.

To address this, we propose \textbf{GenWorld}, a large-scale high-quality AI-generated video detection dataset with a focus on real scenarios including autonomous driving, indoor navigation, and robot manipulation.
As shown in the left part of Figure~\ref{fig:teaser}, GenWorld features the following characteristics: (1) \textbf{Real-world Simulation}: GenWorld focuses on building a video dataset with distributions similar to real-world scenarios, since generated videos that closely mirror real-world videos have the most profound influence.
(2) \textbf{High Quality}: By carefully designing the prompt construction pipeline and selecting state-of-the-art video generation models, GenWorld encompasses high-quality generated video data, which holds significant discriminative value. 
(3) \textbf{Cross-prompt Diversity}: GenWorld includes 10 different generators that take as inputs different modalities of inputs, including text-to-video, image-to-video, and video-to-video models.
They generate videos with varying degrees of forged information, facilitating in-depth research into the characteristics of different levels of forgery.

We conduct a comprehensive evaluation of state-of-the-art AI-generated video detectors~\cite{chen2024demamba, NPR, stil, f3net} on our dataset. 
The results reveal that existing methods fail to discriminate high-quality generated videos from state-of-the-art world models (e.g., Cosmos~\cite{agarwal2025cosmos}).
This is because they mainly focus on detecting generation flaws in pixel space, but high-quality videos often exhibit high levels of forgery, making them harder to detect. 
While the video generation quality is rapidly and constantly improving, it is still challenging to generate physically plausible videos (e.g., 3D consistency, physical laws). 
Motivated by this, we conduct an in-depth analysis of multi-view consistency in videos with the stereo reconstruction model~\cite{wang2024dust3r}, revealing significant differences between generated and real-world videos. 
We then propose SpannDetector to leverages multi-view consistency priors for detecting AI-generated videos. 
Specifically, SpannDetector integrates stereo reconstruction models with temporal memory to more effectively process video information. 
Additionally, we use an authenticity scorer to score the stereo features and globally average them to determine whether the video is AI-generated.
Experimental results demonstrate that SpannDetector outperforms existing methods by considering 3D consistency with a simple design, emphasizing the potential of incorporating physical plausibility into AI-generated video detection.

\input{table/dataset_summary}

%% file: table/dataset_summary.tex
\begin{table*}[t] \small
\centering
\caption{\textbf{Statistics of real and generated videos in the GenWorld dataset.}}
\vspace{-3mm}
\setlength{\tabcolsep}{8pt}
\begin{tabular}{l|c|ccccc|c|c|c}
\toprule
\textbf{Video Source} & \textbf{Type} &\textbf{Task}  & \textbf{Time} & \textbf{Resolution} & \textbf{FPS} & \textbf{Length} & \textbf{Training Set} & \textbf{Testing Set} & \textbf{Total} \\
\midrule
Kinetics-400~\cite{kay2017kinetics} &\multirow4{*}{Real}  & - &17.05&224-340&-&5-10s& 4800 & 1200 & 6000 \\
Nuscenes~\cite{caesar2020nuscenes} & & -& 19.03 & 900-1600& 12& 20s& 680 & 170 & 850 \\
RT-1~\cite{brohan2022rt1} & & -& 22.12 & 256-320 & 10& 2-3s& 1600 & 400 & 2000 \\
DL3DV-10K~\cite{ling2024dl3dv} & & -& 23.12 & 960-540& 30& 3-10s& 1600 & 400 & 2000 \\
\midrule
Opensora-T~\cite{zheng2024opensora}&\multirow{5}{*}{Fake}  &T2V&24.03&512$\times$512&8&2s& 5236 & 1309 & 6545 \\
Opensora-I~\cite{zheng2024opensora}& &I2V&24.03&512$\times$512&8&2s& 5253 & 1314 & 6567 \\
Latte~\citep{ma2024latte} & &T2V&24.03&512$\times$512&8&2s& 7880 & 1970 & 9850 \\
SEINE~\citep{chen2023seine} & &I2V&24.04&1024$\times$576&8& 2-4s& 7880 & 1970 & 9850 \\
ZeroScope~\citep{zeroscopexl} & &T2V&23.07&1024$\times$576&8& 3s& 7880 & 1970 & 9850 \\
\midrule
ModelScope~\citep{wang2023modelscope} &\multirow{5}{*}{Fake}  &T2V&23.03&256$\times$256&8&4s& 7880 & 1970 & 9850 \\
VideoCrafter~\citep{chen2024videocrafter2} &  &T2V&24.01&1024$\times$576&8& 2s& 7880 & 1970 & 9850 \\
HotShot~\citep{Hotshot} & &T2V&23.10&672$\times$384&8& 1s& 7880 & 1970 & 9850 \\
Lavie~\citep{Lavie} & &T2V&23.09&1280$\times$2048&8&2s& 7880 & 1970 & 9850 \\
Cosmos~\cite{agarwal2025cosmos}& &V2V&25.01&640$\times$1024&8&1-5s& 5907 & 1477 & 7384 \\
\midrule
\textbf{Total Count} & - &-&-&-&-&- & 80236 & 20060 & 100296 \\
\bottomrule
\end{tabular}
\label{tab:dataset}
\vspace{-6mm}
\end{table*}

%% file: sec/2_related_work.tex
\section{Related Work}
\textbf{AI-generated Video Dataset.}
AI-generated videos have raised significant concerns due to their potential misuse in telecom fraud and defamatory rumors~\cite{yu2021surveyconcern,golda2024privacyconcern,westerlund2019emergencedeepfakeconcern}. Driven by the powerful facial generation capabilities of GANs~\cite{goodfellow2020generativeGAN} and VAEs~\cite{kingma2013autoVAE}, previous AI-generated video datasets, such as DFDC~\cite{dolhansky2020deepfakeDFDCdataset}, FaceForensics++\cite{rossler2019faceforensics++}, and DFD\cite{dfddataset2019}, primarily focus on deepfake detection. 
However, with the rapid advancement of diffusion models~\cite{ho2020denoisingddpm}, AI-generated forgeries extend beyond facial manipulation~\cite{agarwal2025cosmos,  chen2024videocrafter2, chen2024dreamcinema, huang2024owl1, wang2023modelscope}, demanding more general AI-generated video datasets. GenVideo~\cite{chen2024demamba} initially compiled a large collection of both real and generated videos, but the dataset lacked coherence, with significant discrepancies between real and fake videos. GenVideoBench~\cite{ni2025genvidbench} addressed this issue by incorporating image-to-video models to construct a more structured dataset. However, both methods ignore a fundamental question: \emph{what kind of videos hold real-world significance and truly require detection?} 
This paper focuses on high-quality, impactful real-world videos and present a high-quality AI-generated video dataset of real-world simulations.

\textbf{AI-generated Content Detection.}
As the dataset development progresses, previous research~\cite{xu2023talldeepfake2method, HCILdeepfake3method, gu2021spatiotemporaldeepfake1method} has primarily focused on deepfake video detection due to the lack of high-quality, general AI-generated video datasets. 
STIL~\cite{stil} focuses on capturing the spatial-temporal inconsistencies in forged videos to detect deepfakes. HCIL~\cite{HCILdeepfake3method} uses contrastive learning to capture both local and global temporal inconsistencies between real and fake videos, enabling more robust deepfake detection. TALL~\cite{xu2023talldeepfake2method} transforms a video clip into a pre-defined layout, preserving temporal and spatial dependencies, which enhances generalization for deepfake detection.
However, they are tailored for deepfake videos involving faces, and certain features may fail to detect general AI-generated videos. GenVideo~\cite{chen2024demamba} introduced a plug-and-play temporal module, DeMamba, for detection, but it primarily serves as a temporal fusion module suited for videos.
We conduct an in-depth analysis of high-quality real-world generated videos and identify the inherent multi-view inconsistencies as subtle signs of forgery.
We introduce SpannDetector to leverage 3D consistency for detecting general AI-generated videos.

\textbf{Real-world Video Generation.}
With the advancement of controllable generative models~\cite{zhang2023addingcontrolnet}, video world models~\cite{wang2024drivedreamer, xiang2024pandora, huang2024owl1, bar2024navigationworldmodel, agarwal2025cosmos} for real-world scene video generation have garnered significant attention due to their wide range of downstream applications. 
Navigation World Model~\cite{bar2024navigationworldmodel} predicts future navigation observations based on past observations and navigation actions, leveraging a Conditional Diffusion Transformer (CDiT). 
Pandora~\cite{xiang2024pandora} and Owl-1~\cite{huang2024owl1} employ a hybrid autoregressive-diffusion model to achieve grounded, long-horizon reasoning for human activity scene videos. Cosmos~\cite{agarwal2025cosmos} stands out with its impressive ability to generate videos across multiple scenes. The emergence of video world models for generating real-world scenario videos has further raised concerns about video generation, as these videos can easily deceive humans and spread misinformation. To this end, we aim to build a high-quality AI-generated video dataset of real-world scenes, which will facilitate future research on AI-generated video detectors.

%% file: sec/3_GenWorld.tex
\section{GenWorld}

\subsection{Motivation of GenWorld}
Despite existing datasets~\cite{ni2025genvidbench, chen2024demamba} collecting a large number of AI-generated videos for AI-generated video detector research, they suffer from two fundamental issues in practice: 1) Noisy and Incoherent Semantic Content: Current datasets contain a mix of diverse and unstructured videos, including anime, game footage, comic-style videos, and other content lacking concrete information, as shown in Figure~\ref{fig:teaser}. This raises doubts about the dataset’s relevance—do these videos truly hold value for forgery detection? 2)  Due to the lack of well-designed prompts and the absence of state-of-the-art models, the generated videos are often of poor quality and easily distinguishable by humans. As a result, models trained on these datasets struggle to detect forgeries produced in real-world applications. 

To address this, our GenWorld re-evaluates the problem of which types of generated videos are most impactful and require detection. We believe the most critical videos for detection should possess the following characteristics: 1) Real-World Simulation: Videos such as comics or abstract content have minimal real-world impact. In contrast, those mimicking real-world scenarios, such as driving scenes or human activities, are more likely to influence reality. 2) High-Quality and Realistic: The more realistic a video appears, the more likely it is to mislead humans and impact the real world. Therefore, AI-generated video detection should prioritize high-quality generated videos, especially challenging corner cases.
\subsection{Collection and Organization of GenWorld}
Given the above analysis, we aim to build a high-quality, real-world AI-generated video detection dataset. Firstly, for real-world data, we analyze real-world scenarios and the current video generation world models~\cite{huang2024owl1, agarwal2025cosmos}, selecting four key scenarios that capture a broad spectrum of human life: driving, indoor and outdoor navigation, embodied intelligence manipulation, and human activities, which we include as the core scenes for our dataset. 
\input{figures/pipeline}

For AI-generated videos, we design a data generation pipeline aimed at producing videos that simulate real-world scenarios, as in Figure~\ref{fig:pipeline}. Specifically, we first leveraged a powerful video understanding model, Video-Llava~\cite{lin2024videollava}, to annotate the content of the selected real-world videos, facilitating downstream generation. During the generation process, various generation methods were employed, including text-to-video~\cite{zheng2024opensora, ma2024latte, zeroscopexl, wang2023modelscope, chen2024videocrafter2, Hotshot, Lavie}, image-to-video~\cite{zheng2024opensora, chen2023seine}, and video-to-video~\cite{agarwal2025cosmos}, as these methods simulate real-world videos in different ways, each offering distinct analytical value. In particular, the text-to-video method forges semantic content while still preserving the model's appearance preferences. The image-to-video method forges both semantics and appearance, resulting in a higher level of video forgery. Furthermore, the video-to-video model not only forges semantics and appearance but also manipulates physical laws, demonstrating the highest level of forgery capability. 

Table~\ref{tab:dataset} presents the statistics of our GenWorld dataset. For real-world videos, we randomly select 10,850 samples from Kinetics-400~\cite{kay2017kinetics}, NuScenes~\cite{caesar2020nuscenes}, RT-1~\cite{brohan2022rt1}, and DL3DV-10K~\cite{ling2024dl3dv}.  For AI-generated videos, we utilized 10 different generation models, including state-of-the-art models like Cosmos~\cite{agarwal2025cosmos} and OpenSora~\cite{zheng2024opensora}, spanning various forgery levels such as text-to-video, image-to-video, and video-to-video, resulting in a total of 89,446 generated videos. Our dataset consists of 100,296 videos, with 20\% allocated for testing and the remaining for training. Due to the adoption of state-of-the-art generation models and prompts derived from real-world scenarios, our dataset offers three key advantages: diverse real-world simulation, high-quality video content, and cross-generator as well as cross-generation methods. This rich composition enables various combinations, allowing for a comprehensive analysis of detection methods across multiple dimensions.

%% file: figures/pipeline.tex
\begin{figure}[t]
    \centering
    \includegraphics[width=1\linewidth]{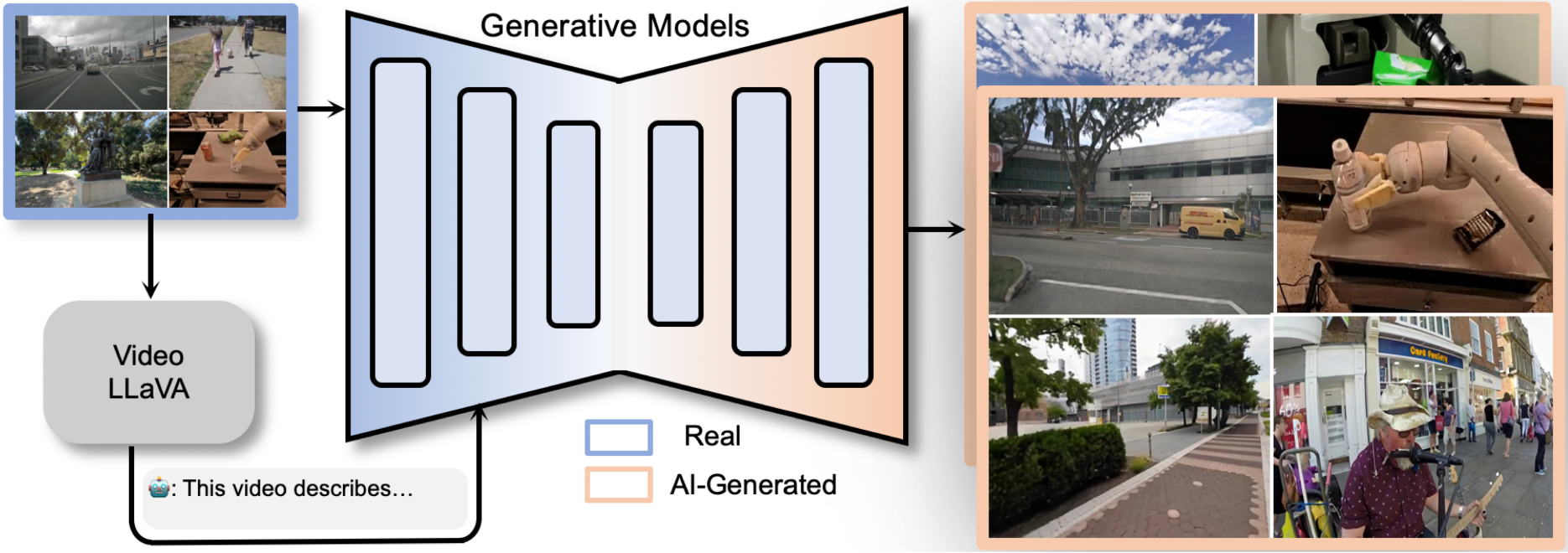}
    \vspace{-7mm}
    \caption{\textbf{Video Generation Pipeline of GenWorld}. }
    \vspace{-5mm}
    \label{fig:pipeline}
\end{figure}

%% file: sec/4_SpannDetector.tex
\section{SpannDetector}
\input{figures/method}

While constructing the dataset, we tested various detection methods~\cite{chen2024demamba, stil, NPR, f3net} and found their performance to be inadequate, especially against the latest high-quality generation models~\cite{agarwal2025cosmos}. This prompted us to explore new perspectives for identifying AI-generated videos. Considering the temporal and multi-view nature of videos, we conducted an in-depth analysis of multi-view consistency and identified it as a potential cue for detecting AI-generated videos (Section~\ref{Method motivation}). Building on this insight, we designed SpannDetector, a multi-view consistency-based AI-generated video detector, which demonstrates promising capabilities, especially against visually hyper-realistic videos produced by world models (Section~\ref{subsec:Spanndetector}). We first review multi-view matching techniques (Section~\ref{subsec:preliminaries}).
\subsection{Preliminaries on Multi-view Matching}
\label{subsec:preliminaries}
Multi-view matching~\cite{wang2024dust3r, wang20243dspann3r} aims to identify a set of corresponding points in multiple views that represent the same 3D structure. The key to success lies in the existence of a 3D structure that satisfies the ill-posed constraints across the multiple views, which can be referred to as multi-view consistency. Recently, Dust3R~\cite{wang2024dust3r}, trained on large-scale data, has demonstrated powerful capabilities in performing multi-view matching quickly and accurately. Formally, given two views of the same 3D structure, \( I^1, I^2 \in \mathbb{R}^{W \times H \times 3} \), Dust3R estimates their point maps in the camera coordinate system of \( I^1 \), denoted as \( X^{1,1}, X^{2,1} \in \mathbb{R}^{W \times H \times 3} \), along with corresponding confidence maps \( C^{1,1}, C^{2,1} \). This can be formulated as:
\begin{equation}
    X^{1,1}, X^{2,1}, C^{1,1}, C^{2,1} = F(I^1, I^2),
\end{equation}
where \( F(\cdot, \cdot) \) denotes the Dust3R model, and \( X^{1,1}, X^{2,1} \) can be used for subsequent camera parameters estimation.

\subsection{Multi-view Consistency of Generated Videos}
\label{Method motivation}
Utilizing the powerful Dust3R, we analyze the multi-view consistency of generated videos. Specifically, for any two frames \( I^1 \) and \( I^2 \) in the video, we first use Dust3R to obtain the corresponding point maps \( X^{1,1} \) and \( X^{2,1} \) and the corresponding camera intrinsic \( K^1 \). We then project \( I^2 \) onto the pixel coordinates of \( I^1 \) with the projection transformation 
\(P^{2,1} = K^1 X^{2,1}\),
where \( P^{2,1} \) represents the 2D projection of the points in \( I^1 \)'s pixel coordinates. This allows us to obtain the projection \( I^{2,1} \) from \( I^2 \) to \( I^1 \). Afterward, we calculate the residual \( R = \left| I^1 - I^{2,1} \right|\), with the result shown in Figure~\ref{fig:method}(b). 
Figure~\ref{fig:method} shows that, for real videos, the residual \(R\) appears quite regular, as real videos inherently maintain multi-view consistency. This allows multi-view matching to easily find a reasonable 3D structure that satisfies the view constraints. However, for AI-generated videos, 
\(R\) is irregular and noisy, indicating that the generated videos exhibit inconsistencies between different views, causing the model to struggle in finding a reasonable 3D structure that simultaneously satisfies the constraints from all views.
\subsection{Design of SpannDetector}
\label{subsec:Spanndetector}
Based on the above analysis, we aim to design an AI-generated video detector that incorporates a multi-view consistency prior. Inspired by Spann3R~\cite{wang20243dspann3r}, we combine a memory module with Dust3R~\cite{wang2024dust3r} to improve detection on sequential video data, as shown in Figure ~\ref{fig:method}. Specifically, when the \( t \)-th frame is input, we use Dust3R Encoder to process both the \( t \) and \( t-1 \) frames simultaneously, obtaining features \( f^{t}_e \) and \( f^{t-1}_e \). Then, we use \( f^{t}_e \) to read the memory information, obtaining \( f^{t}_c \). Afterward, \( f^{t}_e \) and \( f^{t}_c \) are input into Dust3R Decoder, yielding the decoded feature \( f^{t}_d \). Finally, the \( f^{t}_d \) feature is processed by an authenticity scorer to obtain \( s^{t} \). The entire process can be expressed as:
\begin{gather}
    f^{t}_e, f^{t-1}_e = F_{\text{enc}}(I^t, I^{t-1}),\\
    f^{t}_c = \text{Memory\_read}(f^{t}_e, f^{t-1}_{m, k}, f^{t-1}_{m, v}),\\
    f^{t}_d = F_{\text{dec}}(f^{t}_e, f^{t}_c),\\
    s^{t} = \text{Scorer}(f^{t}_d),
\end{gather}
where \( F_{\text{enc}} \) and \( F_{\text{dec}} \) represent the frozen Dust3R's Encoder and Decoder, and Memory\_read is an attention operation. \( f^{t-1}_{m, k} \), and \( f^{t-1}_{m, v} \) are the memory parameters.
At the same time, we use the information from the \( t \) frame to update the memory information as follows:
\begin{equation}
    f^{t}_{m, k}, f^{t}_{m, v} = \Phi(f^{t}_d, s^{t}, f^{t-1}_{m, k}, f^{t-1}_{m, v}).
\end{equation}
After obtaining all the score features \( {\{s^{t}\}^T_{t=0}}\), we perform global average to obtain the final video score, which indicates whether the video is real or fake.

%% file: figures/method.tex
\begin{figure*}[t]
    \centering
    \includegraphics[width=1\linewidth]{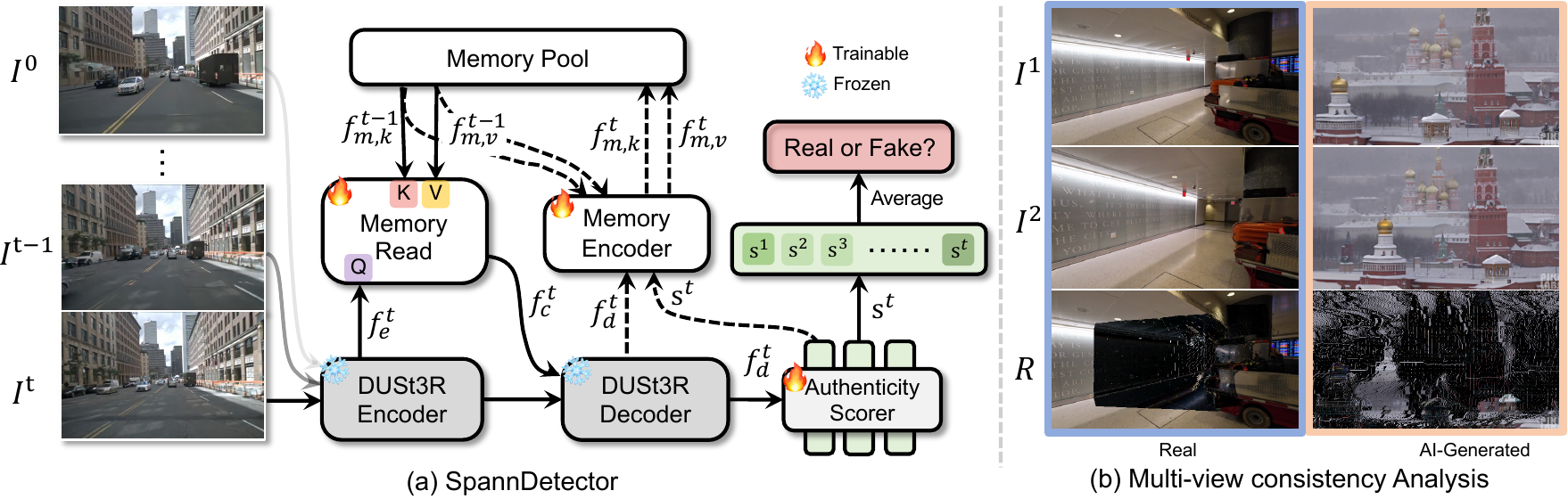}
    \vspace{-7mm}
    \caption{\textbf{Pipeline and motivation of Our SpannDetector.} SpannDetector is designed based on an in-depth analysis of multi-view consistency in real and AI-generated videos. It integrates a stereo reconstruction model with a temporal memory module to enhance efficiency in consistency detection. An authenticity scorer evaluates the stereo features, and the final video authenticity is determined by averaging these scores across the entire video.}
    \label{fig:method}
    \vspace{-6mm}
\end{figure*}

%% file: sec/5_experiments.tex
\section{Experiments}
\subsection{Implementation Details}
\textbf{Datasets.}
To provide a comprehensive evaluation of existing methods, we designed two tasks and partitioned the dataset accordingly. For each model, $20\%$ of the generated data is reserved as the test set. During the evaluation, training is conducted using the training sets from one model, while testing is done using the test sets from other models.
The first task, \textit{Train-Test Evaluation}, simulates a real-world scenario where images of unknown origin must be identified. This is a common situation since we have a limited number of generative models, and the images may come from previously unseen models. In this task, we train the methods using the training sets from five models (Opensora-T~\cite{zheng2024opensora}, Opensora-I~\cite{zheng2024opensora}, Latte~\cite{ma2024latte}, SEINE~\cite{chen2023seine}, and ZeroScope~\cite{zeroscopexl}) as well as real data. We then test the models using the test sets from five other models (HotShot~\cite{Hotshot}, Cosmos~\cite{agarwal2025cosmos}, ModelScope~\cite{wang2023modelscope}, Lavie~\cite{Lavie}, and VideoCrafter~\cite{chen2024videocrafter2}), as well as real data.
The second task, \textit{Cross-prompt Evaluation}, tests how well detection methods perform across different levels of forgery without overfitting to a particular type of generation. In this task, we split the data into T2V, I2V, and V2V generated videos. Models are trained on one of these datasets, and then tested on the test sets from the other two.

\input{table/dataset_comparison}
\input{figures/visualize}

\input{figures/visualize-2}

\textbf{Evaluation Metrics.}
Consistent with previous research~\cite{chen2024demamba, ni2025genvidbench}, we use Accuracy (Acc.) to evaluate the effectiveness of our methods, with AP, F1, and Recall (R) as supplementary evaluation metrics. For image-level detection methods, we combine the predictions of all frames to obtain an overall result. It is important to note that, when calculating Accuracy, we use the test data from the generative model itself to assess the model's ability to differentiate content generated by that specific model. For AP, F1, and Recall, we incorporate the real video test set to ensure a more comprehensive and accurate evaluation.

\textbf{Baseline.}
We select state-of-the-art methods in AI-generated content detection, including both image~\cite{NPR, f3net} and video~\cite{stil,xu2023talldeepfake2method, tong2022videomae, chen2024demamba} detectors, and evaluate their performance. All models are trained on a single A6000 GPU.

\subsection{Comparison with Existing Datasets}
Table~\ref{tab:datasets_Comparison} compares our dataset with existing AI-generated video detection datasets. We highlight the following advantages: 1) Real-World Simulation: Our real videos are carefully selected from diverse datasets that cover most real-world scenarios and derived from these real videos. 
As a result, it lays a solid foundation for developing AI-generated video detection methods with real-world applicability. 2) Cross-prompt diversity: Generation methods from different prompt manipulate real videos to varying extents. Text-to-video generation primarily imitates the semantics and content of real videos while retaining the stylistic preferences in appearance. Image-to-video generation preserves both appearance and semantics while generating highly deceptive videos. Furthermore, video-to-video generation replicates appearance and semantics and simulates the temporal evolution of videos, making them even more temporally realistic. Our GenWorld dataset is the first to encompass text-to-video, image-to-video, and video-to-video generation, paving the way for comprehensive research into the forgery characteristics of these diverse methods. Figure~\ref{fig:visualize} presents examples from our dataset, showcasing its diversity, rich content, and high quality. Meanwhile, Figure~\ref{fig:visualize-temporal} shows the temporal coherence of the videos, highlighting the smooth motion and logical progression over time.

\input{table/Train-Test-Evaluation}

\input{table/F1AP}
\subsection{Train-Test Evaluation}
\vspace{-2mm}
Table~\ref{tab:train-test-evaluation} shows the performance of different models on the Train-Test Evaluation task. Several key observations can be made: 1) Video level AI-generated video detectors significantly outperform image-based detectors. This is primarily because video-based detectors capture temporal information from different time frames, enabling richer representations for better forgery detection. 2) In video level detectors, DeMamba~\cite{chen2024demamba} and VideoMAE~\cite{tong2022videomae} show significantly better performance than STIL~\cite{stil} and TALL~\cite{xu2023talldeepfake2method}. This is because the latter two are primarily designed for deepfake video detection, whereas the formers are designed to capture more general video features. This prompts us to recognize that methods effective for deepfake detection may not necessarily work for general AI-generated videos. Thus, more in-depth research is needed to develop more effective detectors for general AI-generated videos. 3) The difficulty of distinguishing videos generated by different models varies. As shown in the table, most of the trained models perform better on the Lavie~\cite{Lavie} and VideoCrafter~\cite{chen2024videocrafter2} test sets, but perform worse on HotShot~\cite{Hotshot}, ModelScope~\cite{wang2023modelscope}, and Cosmos~\cite{agarwal2025cosmos}. This indicates that the videos generated by the latter are more challenging to detect. Additionally, it is worth noting that videos generated by Cosmos show significantly higher detection difficulty compared to those from other models. This suggests that the world-model-based Cosmos generates videos that are much closer to real-world footage. 4) Our model outperforms others overall, with a significant improvement in performance on detecting videos generated by Cosmos. This is because, in addition to considering temporal dynamics, our model integrates a multi-view consistency prior, which allows it to identify videos that seem realistic but subtly violate physical laws. This highlights the potential of using physical priors as a promising approach for detecting AI-generated videos.

Table~\ref{tab:F1AP} provides the recall (R), F1 score, and average precision (AP) for a more detailed comparison. From the table, we can observe that our model significantly outperforms others in terms of recall (R), while maintaining a high AP, resulting in the highest F1 score. Other models, such as DeMamba, exhibit a relatively low recall (R), particularly when tested on Cosmos-generated videos. This further emphasizes the challenge of detecting Cosmos, as it produces highly realistic real-world simulations that are difficult to distinguish from authentic videos.

\subsection{Cross-prompt Evaluation}
\input{table/cross-generation-method-evaluation}
We also select the representative model DeMamba for Cross-prompt Evaluation, and the results are shown in Table~\ref{tab:cross-evaluation}. As observed in the table, DeMamba tends to overfit the training data and fails to distinguish the data generated from the other two prompts. This suggests that different generation prompt produce distinct artifact patterns, leading the model to learn only those specific features rather than generalizable ones. Moreover, DeMamba exhibit overfitting when trained on all the three cross-prompt data, indicating that this model is not well-suited for learning general forgery detection features. In contrast, our model, after being trained on data generated from one prompt, demonstrates a certain degree of discriminative ability across data generated from other prompts. This suggests that our model may have captured more generalizable features, such as multi-view inconsistency.

\subsection{User Study}
To visually compare the quality and significance of different datasets, we have also conducted a user study to evaluate both existing datasets and our own. The evaluation metrics include Video Quality (VQ), Real-world Simulation (RS), Motion Coherence (MC), and Physical Plausibility (PP). We selected 120 participants from different age groups and randomly chose 100 videos from each dataset for them to evaluate using a 7-point Likert scale. Additionally, each participant was asked to guess whether each video was a real video or an AI-generated video. The experimental results are shown in Table~\ref{tab:user_study}. The experimental results demonstrate that our dataset outperforms the other two across all quality metrics. Furthermore, the user accuracy in determining whether a video is real or AI-generated is lowest for our dataset, highlighting that the generated videos are the most deceptive, 
This makes our dataset particularly valuable for AI-generated video detection research.

\input{table/User_Study}

%% file: table/dataset_comparison.tex
\begin{table*}[t] \small
\centering
\caption{\textbf{Comparison with Existing AI-generated Datasets.} An overview of fake video detection datasets. The proposed GenWorld is the first dataset with a scale of 100,000 containing real-world simulation used to generate videos. Furthermore, GenWorld outperforms other datasets as it incorporates a cross-prompt diversity, including text-to-video, image-to-video, and video-to-video generation. This enables comprehensive analyses of generation methods based on different prompts and their respective levels of forgery.}
\vspace{-3mm}
\setlength{\tabcolsep}{12pt}
 \begin{tabular}{c|c|c|c|c|c|c}
 \toprule
       Dataset& \textbf{Scale} & \textbf{P/Image} &\textbf{P/Video} & \textbf{Semantic Label} & \textbf{Cross Source} &\textbf{Real-world Sim.}\\
 \midrule
   GVD~\cite{bai2024aiGVD}   & 11k   & $\times$   & $\times$     & $\times$ & $\times$ & $\times$\\
   GVF~\cite{ma2024decofGVF}   & 2.8k  & $\surd $   & $\times$      & $\surd $  & $\times$ &$\times$\\
   GenVideo~\cite{chen2024demamba} & 2271k & $\times$     & $\times$    & $\times$ & $\times$ &$\times$\\
   GenVidBench~\cite{ni2025genvidbench} & 143k  & $\surd $      & $\times $      & $\surd $ & $\surd $ & $\times$\\
   \midrule 
   GenWorld &100k & $\surd$ & $\surd$ & $\surd$ & $\surd$ & $\surd$\\
 \bottomrule
 \end{tabular}
\vspace{-2mm}
\label{tab:datasets_Comparison}
\end{table*}

%% file: figures/visualize.tex
\begin{figure*}[t]
    \centering
    \includegraphics[width=0.96\linewidth]{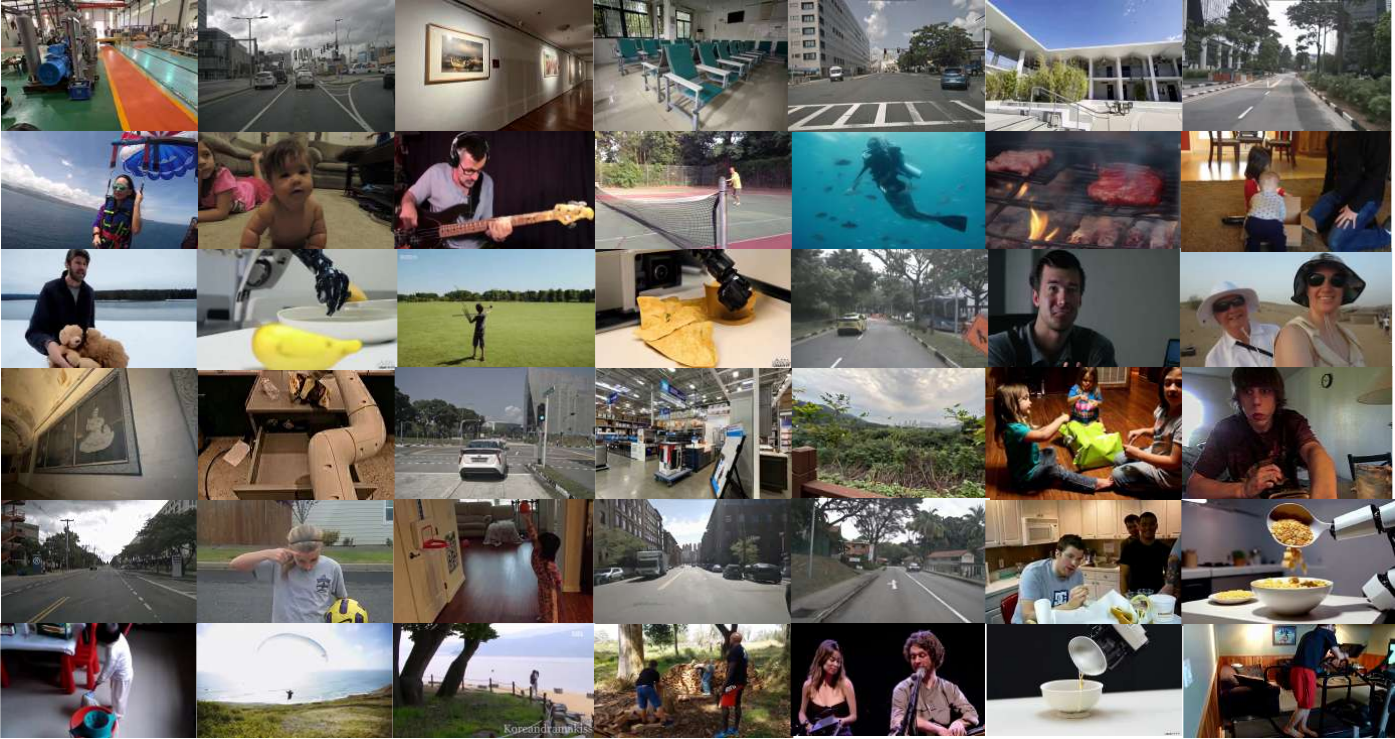}
\vspace{-3mm}
    \caption{\textbf{More Visualization Results of our dataset.} }
    \vspace{-6mm}
    \label{fig:visualize}
\end{figure*}

%% file: figures/visualize-2.tex
\begin{figure*}[t]
    \centering
    \includegraphics[width=1\linewidth]{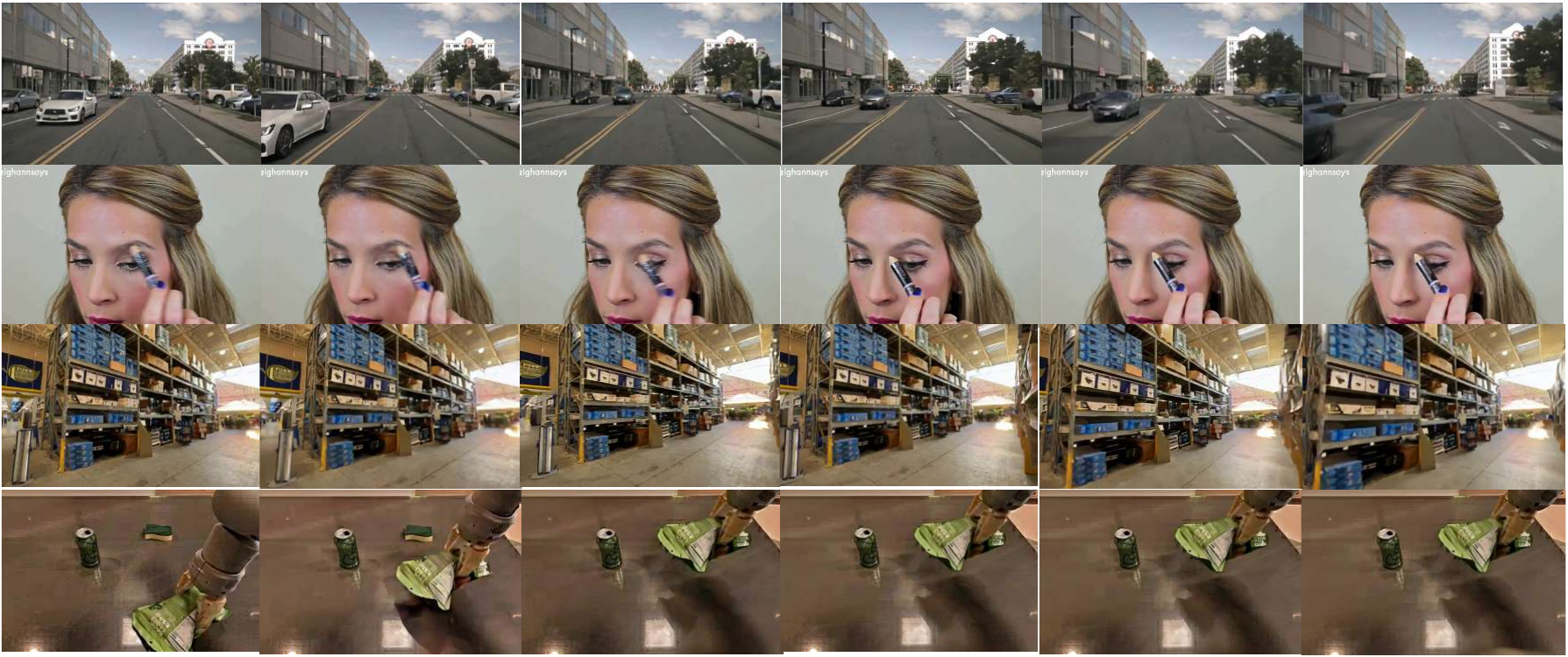}
\vspace{-7mm}
    \caption{\textbf{Temporal Visualization Results of our dataset.} }
    \vspace{-4mm}
    \label{fig:visualize-temporal}
\end{figure*}

%% file: table/Train-Test-Evaluation.tex
\begin{table*}[t] \small
    \centering
    \caption{\textbf{Comparisons to the SOTAs in AI-generated video detection accuracy}
\vspace{-3mm}
    \label{tab:train-test-evaluation}}
\setlength{\tabcolsep}{8pt}
    \begin{tabular}{c | c | c c c c c c}
        \toprule
        Method & Det. & $\text{HotShot~\cite{Hotshot}}$ & $\text{Cosmos~\cite{agarwal2025cosmos}}$ & $\text{ModelScope~\cite{wang2023modelscope}}$ & \text{Lavie~\cite{Lavie}} &\text{VideoCrafter~\cite{chen2024videocrafter2}} &$\text{Average}$ \\
        \midrule 
        F3Net~\cite{f3net} & Imgs & 63.25 & 59.61 & 58.22 & 30.91 &55.04 &52.22 \\
        NPR~\cite{NPR} & Imgs & 34.21 & 29.57 & 71.83 & 17.06 &22.79 &42.17 \\
        STIL~\cite{stil} & Vids &33.15 &27.62 &29.09 &51.78 &53.76 &44.49\\
        TALL~\cite{xu2023talldeepfake2method}&Vids&81.00&44.51&75.83&71.32&91.22&76.24\\
        VideoMAE~\cite{tong2022videomae}&Vids&\textbf{97.56}&30.38&\textbf{93.15}&81.02&99.59&85.40\\
        DeMamba~\cite{chen2024demamba} & Vids & 83.10 & 16.17 & 82.28 & \textbf{99.49} &\textbf{99.95} &82.87\\
         \midrule 
        \textit{Ours} & Vids & 96.24 & \textbf{72.44} & 86.04 & 95.79 & 99.64 &\textbf{89.06} \\
        \bottomrule
        \end{tabular}
\vspace{-6mm}
\end{table*}

%% file: table/F1AP.tex
\begin{table*}[t]
    \centering
    \caption{\textbf{Comparisons to the SOTAs in F1 score (F1) and average precision (AP) on the Train-Test Evaluation.}}
\vspace{-3mm}
    \label{tab:F1AP}
    \resizebox{\textwidth}{!}{
    \begin{tabular}{ccccccccc}
    \toprule
    Model&Det. Lv.&Metric&HotShot~\cite{Hotshot}&Cosmos~\cite{agarwal2025cosmos}&ModelScope~\cite{wang2023modelscope}&Lavie~\cite{Lavie}&VideoCrafter~\cite{chen2024videocrafter2}&Avg.\\
       \midrule
       \multirow{3}{*}{F3Net~\cite{f3net}}&\multirow{3}{*}{Image}&R
        & 0.6325& 0.5961 &0.5822 &0.3091&0.5574 &0.3657\\
        & &F1&0.5862&0.5192&0.5525&0.3370&0.5353&0.4904\\
        & &AP& 0.5878& 0.4842& 0.5392&0.4176&0.5329& 0.6317\\
        \midrule
        \multirow{3}{*}{NPR~\cite{NPR}}&\multirow{3}{*}{Image}&R
        & 0.3421& 0.2957 &0.7183 &0.1706&0.2279 &0.3657\\
        & &F1&0.4282&0.3613&0.7277&0.2391&0.3072&0.4904\\
        & &AP& 0.5516& 0.4403& 0.8100&0.4363&0.4858& 0.6317\\
        \midrule
        \multirow{3}{*}{STIL~\cite{stil}}&\multirow{3}{*}{Videos}&R
        & 0.3315& 0.2760 &0.2909 &0.5178&0.5376 &0.3657\\
        & &F1&0.4000&0.3229&0.3598&0.5617&0.5770&0.4904\\
        & &AP& 0.4973& 0.3978& 0.5062&0.6266&0.6431& 0.6317\\
        \midrule
        \multirow{3}{*}{TALL~\cite{xu2023talldeepfake2method}}&\multirow{3}{*}{Videos}&R
        & 0.8096& 0.4452 &0.7579 &0.7132&0.9122 &0.7425\\
        & &F1&0.8296&0.5445&0.7978&0.7686&0.8878&0.8378\\
        & &AP& 0.8865& 0.6652& 0.8727&0.8288&0.9039& 0.9611\\
        \midrule
        \multirow{3}{*}{VideoMAE~\cite{tong2022videomae}}&\multirow{3}{*}{Videos}&R
        & 0.9756& 0.3038 &0.9315 &0.8102&0.9954 &0.8296\\
        & &F1&0.9732&0.4524&0.9500&0.8808&0.9835&0.9037\\
        & &AP& 0.9928& 0.8116& 0.9854&0.9734&0.9944& 0.9925\\

        \midrule
        
        \multirow{3}{*}{DeMamba~\cite{chen2024demamba}}&\multirow{3}{*}{Videos}&R
        & 0.8310& 0.1617 &0.8228 &0.9949&0.9995 &0.7935\\
        & &F1&0.9057&0.2771&0.9008&0.9954&0.9977&0.8844\\
        & &AP& 0.9919& 0.7300& 0.9944&0.9999&1.0000& \textbf{0.9988}\\

        \midrule
        \multirow{3}{*}
        {Ours}&\multirow{3}{*}{Videos}&R
        & 0.9624& 0.7240 &0.8604 &0.9579&0.9964 &\textbf{0.9095}\\
        & &F1&0.8906&0.7281&0.8356&0.8882&0.9077&\textbf{0.9322}\\
        & &AP& 0.9475& 0.7606& 0.9062&0.9446&0.9777& 0.9560\\
        \bottomrule
    \end{tabular}}
\vspace{-5mm}
    \label{main_results}
\end{table*}

%% file: table/cross-generation-method-evaluation.tex
\begin{table}[t] \small
    \centering
    \caption{\textbf{Results of our cross-prompt evaluation.} \label{tab:cross-evaluation}}
\vspace{-3mm}
\setlength{\tabcolsep}{6pt}
    \begin{tabular}{c | c | c c c c}
        \toprule
        Method & $\text{Gen-M}$ & T2V & I2V & V2V &Avg.\\
        \midrule
    \multirow3{*}{DeMamba~\cite{chen2024demamba}}& T2V & 98.41 & 9.35 & 0.61&75.90\\
         & I2V & 30.33 &100 &0.27&46.95\\
         & V2V &53.54 & 4.93 & 100&47.27 \\
        \midrule 
         \multirow3{*}{Ours} &T2V &99.05 &51.66 &52.37&86.09\\
         & I2V &52.47 &98.60 &50.34&64.38\\
         & V2V &59.89 &28.71 &96.35&60.21\\
          \bottomrule
        \end{tabular}
\vspace{-5mm}
\end{table}

%% file: table/User_Study.tex
\begin{table}[t!] \small
    \centering
    \caption{\textbf{User study on the dataset quality.} \label{tab:user_study}}
\vspace{-3mm}
\setlength{\tabcolsep}{8pt}
    \begin{tabular}{c | c c c}
        \toprule
        Dataset & GenVideo~\cite{chen2024demamba} & GenVidBench~\cite{ni2025genvidbench} & Ours \\
        \midrule
        VQ&5.4&5.9&6.3\\
        RS&3.2&4.4&6.4\\
        MC&5.2&5.8&6.2\\
        PP&5.1&5.3&6.4\\
        Acc.&100\%&98.6\%&89.4\%\\
          \bottomrule
        \end{tabular}
\vspace{-6mm}
\end{table}

%% file: sec/6_conclusion.tex
\section{Conclusion}
In this paper, we introduce GenWorld, a high-quality dataset for detecting AI-generated videos that simulate real-world scenarios. GenWorld is characterized by three key aspects: 1) \textbf{Real-World Simulation:} It focuses on videos that closely mimic real-world activities, potentially having a greater impact on societal events. 2) \textbf{High Quality:} By leveraging multiple state-of-the-art generation methods, GenWorld comprises a large collection of high-quality AI-generated videos. 3) \textbf{Cross-prompt Diversity:} It includes videos generated from diverse prompts—text, images, and videos—capturing varying levels of forgery characteristics. We conduct an in-depth evaluation of existing advanced AI-generated video detectors using GenWorld and find that they struggle to detect high-quality videos generated by world models (e.g., Cosmos~\cite{agarwal2025cosmos}). 
This limitation likely stems from their failure to capture real-world physical consistency. To address this, we propose SpannDetector, a simple yet effective method built on an in-depth analysis of 3D consistency between real and generated videos. SpannDetector integrates a stereo reconstruction model with a temporal memory module, significantly improving detection performance, particularly on world model-generated videos, which highlights the potential of leveraging physical consistency for AI-generated video detection.